\documentclass[a4paper]{svproc}

\usepackage{url}

\usepackage{hyperref}

\usepackage{graphicx}
\graphicspath{{figures/}}
\usepackage{capt-of}
\usepackage{varwidth}

\usepackage{booktabs}
\usepackage{multirow}

\usepackage{color}
\usepackage{xspace}
\usepackage{adjustbox}

\usepackage{amsmath}
\usepackage{amsfonts}
\usepackage{newtxmath}
\DeclareMathOperator*{\argmax}{argmax}

\usepackage{autolabtools}

\newcommand{\algname}{Distribution Area Reduction for Stacked Scenes\xspace}
\newcommand{\algabbr}{DARSS\xspace}

\newcommand{\SR}{SR}
\newcommand{\SA}{Steps}

\newcommand{\sr}[1]{\ifthenelse{\equal{#1}{*}}{\srStar}{\srNoStar{#1}}}
\newcommand{\srStar}[1]{\textbf{\srNoStar{#1}}}
\newcommand{\srNoStar}[1]{#1\%}

\newcommand{\sa}[1]{\ifthenelse{\equal{#1}{*}}{\saStar}{\saNoStar{#1}}}
\newcommand{\saStar}[3]{\textbf{\saNoStar{#1}{#2}{#3}}}
\newcommand{\saNoStar}[3]{#1 (#2 -- #3)}

\renewcommand{\remark}[3]{\marginpar{\scriptsize\textcolor{#2}{#1: #3}}}

\definecolor{orange}{rgb}{0.53, 0.53, 0.19}

\definecolor{britishracinggreen}{rgb}{0.23, 0.53, 0.19}
\newcommand{\jeffi}[1]{\remark{JI}{red}{#1}}
\definecolor{turquoise}{rgb}{0.25, 0.88, 0.82}

\definecolor{navy}{rgb}{0,0,0.5}

\renewcommand{\remark}[3]{}

\usepackage[backend=biber,
            hyperref=true,
            url=false,
            isbn=false,
            doi=false,
            backref=false,
            natbib=true, 
            mincitenames=1,
            maxcitenames=3,
            citestyle=numeric-comp,
            sorting=none,
            block=none]{biblatex}

\AtBeginBibliography{\setcounter{maxnames}{500}}

\begin{document}
\mainmatter

\title{Mechanical Search on Shelves with Efficient Stacking and Destacking of Objects}

\titlerunning{Mechanical Search on Shelves with Stacked Objects}

\author{Huang Huang*\inst{1} \and Letian Fu*\inst{1} \and  Michael Danielczuk\inst{1} \and Chung Min Kim\inst{1} \and \\Zachary Tam\inst{1}\and  Jeffrey Ichnowski\inst{1}\and Anelia Angelova\inst{2} \and Brian Ichter\inst{2} \and \\ Ken Goldberg \inst{1} {\scriptsize *equal contribution}}
\authorrunning{Huang et al.}
\tocauthor{Huang Huang, Letian Fu, Michael Danielczuk, Chung Min Kim, Zachary Tam, Jeffrey Ichnowski, Anelia Angelova, Brian Ichter, Ken Goldberg}
\institute{University of California Berkeley, Berkeley CA 94720, USA,\\
\and
Google, USA\\
\email{huangr@berkeley.edu}
}

\maketitle

\begin{abstract}
Stacking increases storage efficiency in shelves, but the lack of visibility and accessibility makes the mechanical search problem of revealing and extracting target objects difficult for robots. In this paper, we extend the lateral-access mechanical search problem to shelves with stacked items and introduce two novel policies---\algname (\algabbr) and Monte Carlo Tree Search for Stacked Scenes (MCTSSS)---that use destacking and restacking actions. MCTSSS improves on prior lookahead policies by considering future states after each potential action. Experiments in 1200 simulated and 18 physical trials with a Fetch robot equipped with a blade and suction cup suggest that destacking and restacking actions can reveal the target object with 82--100\% success in simulation and 66--100\% in physical experiments, and are critical for searching densely packed shelves. In the simulation experiments, both policies outperform a baseline and achieve similar success rates but take more steps compared with an oracle policy that has full state information. In simulation and physical experiments, DARSS outperforms MCTSSS in median number of steps to reveal the target, but MCTSSS has a higher success rate in physical experiments, suggesting robustness to perception noise. See \url{https://sites.google.com/berkeley.edu/stax-ray} for supplementary material.


\end{abstract}
\section{Introduction} \label{sec:introduction}

Finding a desired object on cluttered shelves with limited visibility is a common task in households, pharmacies, and warehouses.
\emph{Mechanical search} aims to reveal a partially or fully occluded target from an environment via interaction with the environment. Prior work in mechanical search has explored both overhead-access bins~\cite{danielczuk2019mechanical, kurenkov2020visuomotor, yang2020deep, zeng2018learning}, where the objects can be heaped in arbitrary poses, and lateral-access environments~\cite{bejjani2020occlusion,gupta2013interactive,huang2020mechanical}, where the objects sit in stable poses. However, these works include strong assumptions, such as no stacked objects~\cite{bejjani2020occlusion,gupta2013interactive,huang2020mechanical,li2016act,wang2021efficient}, availability of space between objects for gripper insertion~\cite{bejjani2020occlusion,gupta2013interactive,li2016act,wang2021efficient}, a partially visible target object~\cite{motoda2021bimanual}, a discretized state space~\cite{gupta2013interactive,wang2021efficient}, or actions that remove objects from the environment~\cite{dogar2014object,zeng2017multi}. In contrast, we consider environments with object stacks, fully occluded target objects, and a continuous state space where all objects must remain within the shelf.

\begin{figure}[t!]
    \centering
    \includegraphics[width=\linewidth]{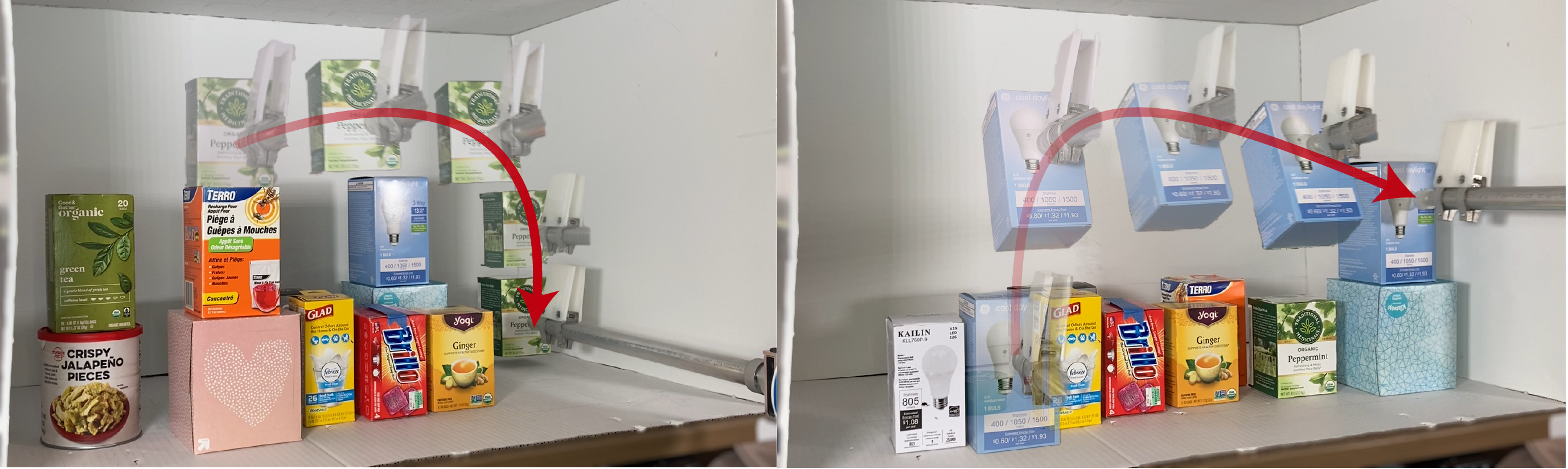}
    \caption{Destacking (\textbf{left}) and restacking (\textbf{right}) actions performed with a bluction end-effector (blade and suction in white and gray; more information in supplementary material) mounted on a Fetch robot; the object trajectory is indicated in red.
    When target objects are hidden behind object stacks, destacking is necessary to reveal them. Restacking helps to open more space without blocking visible objects. 
    DARSS and MCTSSS leverage both action types to reveal target objects in densely packed shelves.}
    \label{fig:splash}
    \vspace{-20pt}
\end{figure}

Stacked objects in shelves are common in home environments, retail stores, and warehouses because they allow for efficient use of storage space, but they introduce novel challenges for mechanical search: stacked objects limit available actions and cannot be reliably moved until the objects are destacked. Restacking objects may be necessary in crowded shelves with limited free space to reveal the target object, but restacking introduces an additional stability requirement: restacked objects must be able to rest stably on the object beneath them. We propose Distribution Area Reduction for Stacked Shelves (DARSS) to reveal target objects in scenes with multiple object stacks. DARSS extends previous lateral access mechanical search work based on occupancy distributions~\cite{huang2020mechanical,huang2022mechanical} by leveraging discrete destacking and restacking actions (Figure~\ref{fig:splash}).

Stacked shelves also motivate mechanical search policies that look ahead to plan actions that do not impede future actions, since destacking is inherently a multi-step process. We extend previous lookahead policies~\cite{huang2020mechanical} that consider only one future state (revealing the back wall of the shelf) by implementing a Monte Carlo Tree Search in the Stacked Scene (MCTSSS) policy that considers future states resulting from three possible outcomes. MCTSSS allows for efficient sampling and evaluation of the exponential number of future actions and states. 

This paper makes three contributions:
\begin{enumerate}
    \item A formulation of the lateral-access mechanical search problem that includes shelves with stacked objects.
    \item Two novel mechanical search policies, Distribution Area Reduction for Stack-ed Scenes (DARSS) and Monte-Carlo Tree Search for Stacked Scenes (MCTS-SS), that include destacking and restacking actions. MCTSSS leverages 3 potential future states after pushing and grasping actions across multiple steps. 
    \item Experiments in simulation and on a physical Fetch robot equipped with a pushing-blade-and-suction-cup (bluction) tool, evaluating the success rate and the number of steps to find a target object using DARSS, MCTSSS, baseline, and oracle policies.
\end{enumerate}

\section{Related Work} \label{sec:relatedwork}
\subsection{Stacking and Destacking}
Stacking and destacking are critical in industrial contexts, such as container unloading~\cite{vaskevicius2014object}. \citet{lee2021beyond} use deep learning for robot stacking, introduce a robot stacking benchmark, and present a vision-based policy using RGB images. It is critical to correctly perceive individual objects in stacks for object extraction and manipulation. 
When object geometries are unknown, their boundaries become unclear, making perception and segmentation difficult. 
\citet{furrer2017autonomous,vaskevicius2014object} use object databases to match object instances and filter unrealistic segmentations. \citet{landgraf2021simstack} propose using multiple views of stacks to predict 3D segmentations. In contrast, we use SD Mask R-CNN~\cite{danielczuk2019segmenting} trained on simulated images with stacked objects to get segmentation masks. \citet{kumar2021graph} learn both scene generation and scene exploration agents.  The generation agent attempts to generate scenes with hidden objects, including scenes with stacks of objects.  The exploration agent attempts to reveal the hidden objects. However, 
they aim to reveal any hidden object instead of guiding exploration toward a target object.

\subsection{Mechanical search}\label{related:ms}
There are numerous methods that tackle variations of the mechanical search~\cite{danielczuk2019mechanical,kurenkov2020visuomotor} problem, or interactive object search~\cite{gupta2012interactive,gupta2013interactive,li2016act,moldovan2014occluded,wong2013manipulation} problem. In these problems, a robot aims to extract a target object by manipulating occluding and distractor objects. Recent work focused both on the overhead-access or tabletop perspective~\cite{danielczuk2020x,kurenkov2020visuomotor,novkovic2020object,yang2020deep,zeng2018learning}, and the lateral-access or shelf perspective~\cite{bejjani2020occlusion,gupta2013interactive,huang2020mechanical,lou2021collision}.

Learning an end-to-end policy to perform mechanical search is difficult. The policy has to learn to operate in continuous action space with different modalities, learn from sparse reward signals, understand object occlusion, and generalize to unseen objects. Moreover, physical mechanical search experiments are expensive to rollout and difficult to reset, making policy learning in real scenes challenging~\cite{danielczuk2019mechanical}. To break up the problem, \citet{danielczuk2019mechanical, huang2020mechanical, huang2022mechanical} use \emph{occupancy distributions} learned from simulated depth images to guide the search policy. The occupancy distribution is a continuous function in image space, where each pixel value represents the probability there is an occluded target object (i.e., the likelihood that the pixel would show part of the target object if there were no occluding objects in the scene). In this work, we also leverage occupancy distributions to guide our search for the target object.

Compared to mechanical search in overhead settings, lateral-access environments put tighter constraints on grasping and motion planning. \citet{lou2021collision} plan 6-DOF grasps using a learned collision-aware reachability predictor that estimates collisions between the robot arm and its surroundings from point cloud observations. \citet{bejjani2020occlusion} use continuous pushing actions to slide objects on shelves to reach and extract a target behind. \citet{zenkri2022hierarchical} formulate the mechanical search problem as a hierarchical POMDP and propose a hierarchical policy learning approach. \citet{gupta2013interactive} discretize object placements within the shelf and uses a multi-step object search algorithm to push or pick-and-place objects to search for the target object. In contrast, we consider a continuous state space and continuous object placements. 
\citet{huang2022mechanical} use a novel bluction tool, a pneumatically-actuated suction cup combined with a pushing blade, that enables both suction grasps and pushing actions in shelves where there is not enough space between objects for a parallel-jaw gripper. They also introduce several lateral-access mechanical search policies and demonstrate that multi-step lookahead and retaining an exploration history are beneficial in shelves where objects may not be removed. In contrast, we consider shelves with stacks of objects that require destacking or restacking actions to maintain space for exploration.
We leverage the bluction tool for suction grasping, which has been effective for grasping in constrained environments~\cite{mahler2019learning, morrison2018cartman,zeng2017multi}, for stacking and destacking actions within shelves.

\subsection{Monte Carlo Tree Search}
Monte Carlo Tree Search (MCTS)~\cite{coulom2006efficient, kocsis2006} has been used in a variety of machine learning contexts such as AlphaGo~\cite{silver2016mastering}. Within robotics, MCTS has been applied to many domains, such as task and motion planning~\cite{ren2021extended}, assembly~\cite{funk2022learn2assemble}, scene understanding~\cite{hampali2021monte}, and mobile manipulation~\cite{kim2020monte}. Each of these requires a balance of exploration with exploitation to efficiently solve problems in these domains.  A similar balance must be achieved during mechanical search in shelves: when there is a high likelihood that a target object lies behind a large stack of objects, we can either attempt exploring low-risk potential locations for the target (exploration) or take multiple steps to fully destack the objects (exploitation) and risk the possibility the target lies elsewhere.

\section{Problem Statement} \label{sec:problem_statement}


\subsection{Overview}
We consider a shelf of known dimensions with $N$ unknown cylindrical and cuboid-al objects $\left\lbrace \mathcal{O}_1, \, \ldots, \, \mathcal{O}_N \right\rbrace$. Object poses are defined with respect to a coordinate frame with the origin at the center of the shelf floor, $y$-axis pointing toward the shelf opening, $z$-axis normal to the plane of the floor of the shelf, and $x$-axis completing the basis. Objects are resting in one of their stable poses with cuboids having their face normals aligned with the $x$- and $y$-axes of the shelf (i.e., there is always a cuboid face parallel to the back wall of the shelf). An object $\mathcal{O}_i$ may be stacked on top of another object $\mathcal{O}_j$ if the projection of $\mathcal{O}_i$ onto the plane of the shelf is entirely contained within the projection of $\mathcal{O}_j$. There is a single target object $\mathcal{O}_T$ of a known shape and color that is initially occluded on the shelf. The robot views the shelf using an RGBD camera and may push or lift and place objects within the shelf using a bluction tool. Objects cannot be removed from the shelf. The objective is to minimize the number of actions required to reveal the hidden target object.

\subsection{Problem Setup}
We model this problem as a finite-horizon Partially Observable Markov Decision Process (POMDP) with states $\mathcal{S}$, observations $\mathcal{Y}$, actions $\mathcal{A}$, and horizon $H$. An episode ends either when $V\%$ of the target object is revealed or when the horizon $H$ is reached. The robot head-mounted camera takes an RGBD image $\mathbf{y}_t \in \mathbb{R}^{w \times h \times 4}$ of the shelf at each timestep $t$, from which, the target object visibility percentage, $v_t$, is determined assuming the access to an algorithm that computes the visibility.
We define a state $\mathbf{s}_t \in \mathcal{S}$ at timestep $t \in \left\lbrace 1, \, \ldots, \, H \right\rbrace$ as the geometries and poses of all objects $\left\lbrace \mathcal{O}_1, \, \ldots, \, \mathcal{O}_N, \mathcal{O}_T\right\rbrace $ in the environment. 

The robot takes actions $\mathbf{a}_t \in \mathcal{A}$, where $\mathcal{A} = \mathcal{A}_p \cup \mathcal{A}_s$ is the union of pushing and suction actions.
Pushing action parameters are $\left(\textbf{q}, d_x \right)$, and suction action parameters are $\left(\textbf{q}, d_x, d_y, d_z \right)$, where $\mathbf{q}$ is the starting pose of the bluction tool and $(d_x, d_y, d_z)$ are signed travel distances along the respective axes.
Suction actions decompose to 4 linear motions: 1) lifting the object along the $z$-axis, 2) pulling the object out of the shelf along the $y$-axis, 3) translating the object along the $x$- and $z$-axes by $\left(d_x, d_z\right)$, and 4) placing the object onto another object or the shelf at depth $d_y$.

\subsection{Assumptions}
We make the following assumptions: 1) The shelf and camera are fixed. 2) The color and geometry of the target object are known and the target visibility can be accurately measured from the camera. 3) All objects are either cylinders or rectangular prisms. All objects rest in one of their stable poses and rectangular prisms are axis-aligned with the shelf. 4) Actions do not inadvertently topple objects or move multiple objects. 5) An object can be directly supported by at most one other object. 6) Object segmentation masks are accurate and sufficient for a policy to identify the spatial relation between observable objects. 
\subsection{Objective}
The policy episode ends when $V\%$ of the target is revealed (Success) or it reaches the maximum number of steps $H$ (Failure). We assign a cost of 1 to each action performed before termination such that the total cost is the number of steps taken before termination. 

Formally, we aim to find a mechanical-search policy $\pi$ that minimizes the total cost to reveal the target object:
$
    \min_\pi \; \mathbb{E}_{\Pi(\pi)} \left[\sum_{t=1}^h \vmathbb{1}_{\left[0, V\right)}(v_t)\right] 
$
where $\Pi(\pi)$ denotes the space of episodes generated by policy $\pi$, $h\leq H$ is the episode length, $\vmathbb{1}_{\left[0, V\right)}(v_t)$ is the indicator function equal to 1 if $v_t < V$ and 0 otherwise.

\section{Methods}
We introduce two novel mechanical-search policies that use pushing or suction for moving objects including destacking and restacking objects. We use the Lateral-Access X-RAY (LAX-RAY) model~\cite{danielczuk2020x,huang2020mechanical} to predict an occupancy distribution of the target given a depth observation and an object aspect ratio. For both policies, LAX-RAY estimates a 1D occupancy distribution of the target object $P_t(x)$, where $x\in[1 \dots w]$ is the pixel location along the image $x$-axis. We encode the history of previous occupancy distributions via the minimum of the current 1D predicted occupancy distribution and the previous encoding: $P'_t(x) = \min \left\lbrace P_t(x), \; P'_{t-1}(x)\right\rbrace$, with $P'_0(x) = P_0(x)$.

\subsection{Occupancy distribution}
Occupancy distribution implies the possibility of each object hiding the target object as explained in Section~\ref{related:ms}. LAX-RAY model is trained on simulated images with the ground-truth occupancy distributions calculated using Minkowski sum for the corresponding target aspect ratios (ratio of object height to object width)~\cite{huang2022mechanical}. More details are in supplementary material. The learned occupancy distribution therefore encodes the geometries of visible objects and the perspective effect of the camera, where larger objects that are placed closer to the camera cause more occlusion than smaller objects placed farther from the camera. Since all objects sit at their stable poses in the lateral access environment, the 2D occupancy distribution can be collapsed into the 1D distribution by summing the 2D distribution along the image y-axis.

\subsection{Stacked Environments}\label{sec:stack}
A stack is defined as a structured arrangement of objects that meets two conditions: 1) all objects must be directly supported by the shelf floor or by a single object, and 2) the cross-sectional area of any given object must not be larger than that of all objects between the shelf and itself. To avoid invalid actions, such as removing the bottom-most object in a stack, we use a tree structure to track the relationship between the shelf and the objects.

\subsection{Action Generation}
\jeffi{Before $\mathcal{A}_p$ and $\mathcal{A}_s$ meant a continuous parameterized state.  Here they seem to mean discrete sets that are functions of a state.  Perhaps $\hat{\mathcal{A}}_p:\mathcal{S}\rightarrow \{\mathcal{A}_p, \mathcal{A}_p\}$, etc.}
As $\mathcal{A}$ is a continuous set, we reduce the complexity of the policy decisions by generating discrete pushing $\mathcal{A}_p$ and suction $\mathcal{A}_s$ actions for each state $s_t$. The pushing action space $\mathcal{A}_p$ contains at most two actions per object: the maximum translation in the positive and negative shelf $x$-axis that the object can be moved without touching other objects or the shelf wall. 

The suction action space $\mathcal{A}_s$ consists of two types of actions. \emph{Rearrangement actions} translate the object only in the $xy$-plane of the shelf. These actions can be executed only on single objects that rest directly on the shelf floor and do not belong to an object stack. For each object with an available collision-free suction grasp, we discretize the shelf $x$-axis into $B$ equal-size bins. We generate a set of $x$-axis offsets $d_x$ from the current object position to the center of each bin and choose the corresponding collision-free object translation $d_y$ closest to the back shelf wall for each offset $d_x$. By placing objects as far back as possible in the shelf, the object occlusion is minimized in image space, thus maximizing scene visibility and reducing the rearrangement actions space. \emph{Stacking and destacking actions} have non-zero translation in the shelf $z$-axis (i.e., the object is raised onto a stack or lowered off of a stack to the shelf surface). Only the top objects on a stack or objects not belonging to a stack are eligible for stacking actions. For stacking actions, object placement positions are chosen from other visible objects satisfying conditions in Section~\ref{sec:stack}.
Objects at the top of a stack are eligible for destacking actions, which are generated in the same way as rearrangement actions, but include the $d_z$ component to move the object to the shelf surface.

Thus, the generated action set at each step is upper bounded by $(2+B)N + N^2 \in O(N^2+BN)$, as there are up to 2 pushing actions and $B$ rearrangement or destacking actions per object and a maximum of $N^2$ total stacking actions.

\subsection{Distribution Area Reduction for Stacked Shelves (DARSS)}
We introduce a novel mechanical-search policy for shelves with stacks of objects: Distribution Area Reduction for Stacked Shelves (DARSS) based on prior work~\cite{huang2020mechanical}, which greedily selects the action that maximally reduces the target occupancy distribution at each time step until the episode terminates. 
The policy selects both pushing $\mathcal{A}_p$ and suction $\mathcal{A}_s$ actions.

At each step $t$, we define the support of the occupancy distribution for object $\mathcal{O}_i$ as $\sigma_t(\mathcal{O}_i) = \sum_{x=l_i}^{r_i} P'_t(x)$, where $l_i$ and $r_i$ are the pixel locations of the left and right edges of object $\mathcal{O}_i$'s segmentation mask in the image $x$ axis. We further define the reduction of support for each object $\mathcal{O}_i$ as
\begin{equation} \label{eq:reduction_of_support}
    \Delta \sigma_t(\mathcal{O}_i) = \sigma_t(\mathcal{O}_i) - \sigma_{t+1}(\mathcal{O}_i),
\end{equation}
where $\sigma_{t+1}(\mathcal{O}_i)$ is calculated using the object segmentation mask after taking the action.
If an action increases the overlap with the occupancy distribution, $\Delta \sigma_t$ may be negative (i.e., DARSS gives negative scores for blocking objects that may need to be moved in the future). DARSS returns a list of actions sorted by decreasing $\Delta \sigma_t{(\cdot)}$ and executes the first feasible action from the list. Thus, DARSS chooses the action that maximally reduces overlap of objects with the occupancy distribution at each step.

\begin{figure}[t!]
    \centering
    \includegraphics[width=0.8\linewidth]{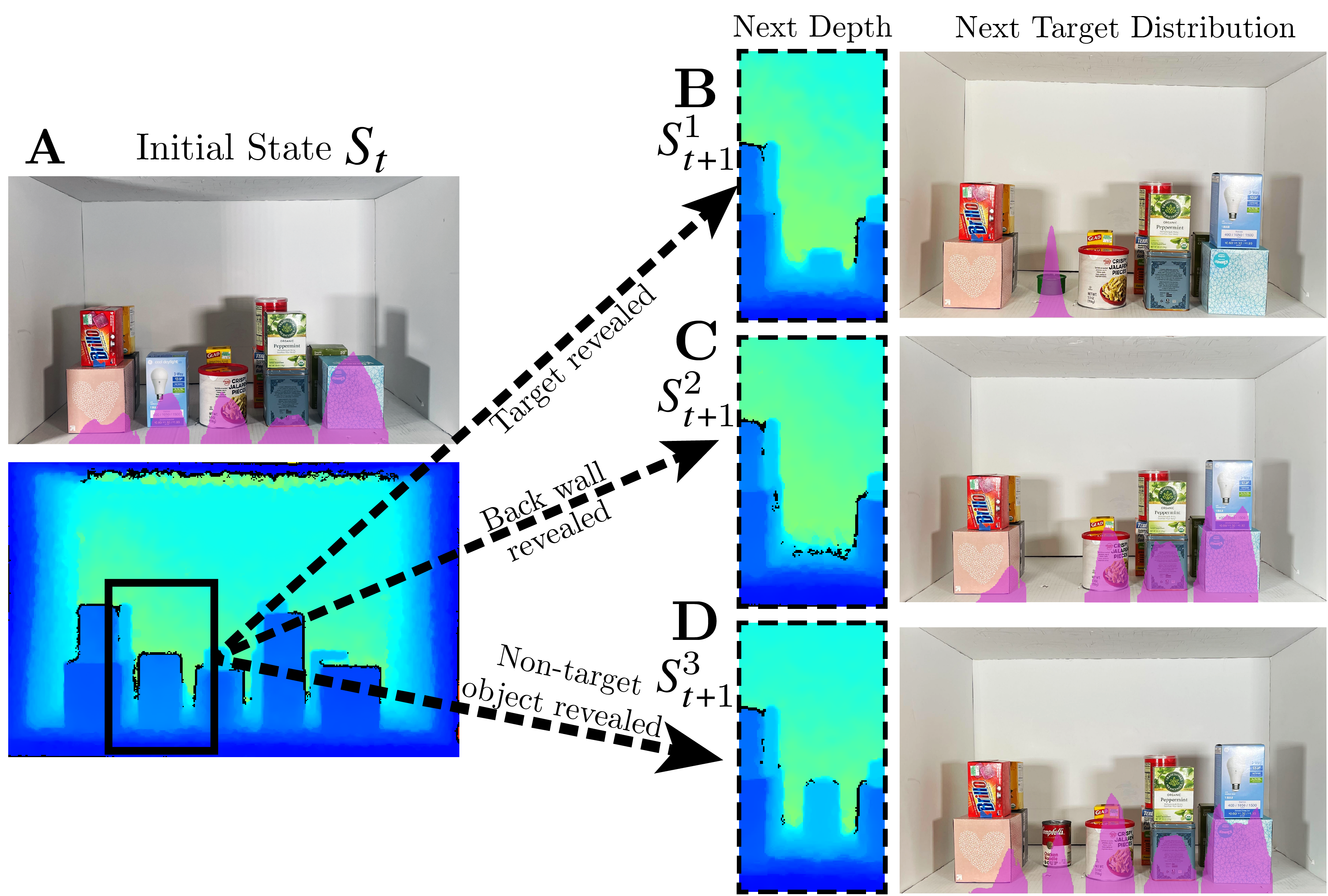}
    \caption{\textbf{A} shows a shelf with a fully-occluded green target object. The RGB image on the top left and depth image on the bottom left are overlaid with the estimated target occupancy distribution in pink. The MCTSSS policy considers 3 possible action outcomes, each with an updated target occupancy distribution: \textbf{B} the target is directly revealed, \textbf{C} the shelf back wall is directly revealed, or \textbf{D} a non-target object is revealed.}
    \label{fig:mctsevol}
    \vspace{-20pt}
\end{figure}

\subsection{Monte Carlo Tree Search for Stacked Shelves (MCTSSS)} 
We build a Monte Carlo Tree Search for Stacked Shelves (MCTSSS) policy by defining a tree of states and actions that grows from root state $S_t$ with state nodes being predicted depth images of pixel size $w{\times}h$ at each tree depth.

In contrast to Distribution Entropy Reduction over $n$ steps (DER-n)~\cite{huang2020mechanical}, a lookahead policy from previous work which always assumes that moving an object will reveal the shelf back wall, MCTSSS considers three possible next states $\left\lbrace S_{t+1}^1, S_{t+1}^2, S_{t+1}^3 \right\rbrace$, with sampling probabilities of $p_1, p_2, p_3$, respectively, as shown in B, C, D of Figure~\ref{fig:mctsevol}: 1) the target is revealed, 2) the shelf back wall is revealed, and 3) an object other than the target is revealed.  MCTSSS approximates the sampling probabilities based on the 1D occupancy distribution for the current observation ${P}'_{t}(x)$:
\begin{align}
    p_1 &= \frac{\sum_{x=1}^w \vmathbb{1}_{\mathcal{O}_i}(x) {P}'_{t,j'}(x) }{\sum_{x=1}^w {P}'_{t,j'}(x)}, \label{eq:state1}\\ 
    p_2 &= 1 - \frac{\sum_{j=0}^2 \sum_{x=1}^w \vmathbb{1}_{\mathcal{O}_i}(x)  {P}'_{t,j}(x) }{\sum_{j=0}^2  \sum_{x=1}^w {P}'_{t,j}(x)}, \label{eq:state2} \\ 
    p_3 &= 1 - p_1 - p_2, \label{eq:state3}
\end{align}
where ${P}'_{t,j}(x)$ is the 1D occupancy distribution for objects with aspect ratio index $j$, $j'$ indicates the target object aspect ratio index, and $\vmathbb{1}_{\mathcal{O}_i}(x)$ indicates that object $\mathcal{O}_i$ overlaps with column $x$ in the image. We approximate all objects as belonging to three aspect ratios, 1:1, 2:1, and 4:1, so $j \in \lbrace 0, 1, 2 \rbrace$\jeffi{The mapping from 1:1, 2:1, and 4:1 to $j \in \{ 0,1,2 \}$ may be confusing to a reader.  It might warrant further explanation/background of $P$}. Intuitively, $p_1$ is the fraction of the occupancy distribution occluded by objects or equivalently how likely the target is to be behind the object to be moved. $p_2$ is a similar probability but takes into account multiple aspect ratios (i.e., 1 minus the likelihood that an object of \textit{any} aspect ratio is behind the object to be moved).


Upon reaching $S_{t+1}^1$, the Monte Carlo rollout terminates with the target revealed. MCTSS evolves states $S_{t+1}^2$ and $S_{t+1}^3$ by generating predicted depth observations after executing the action $\textbf{a}_t$. 
$S_{t+1}^2$ is generated by updating the depth observation at $S_{t}$ through reprojecting the moved object into the image and replacing the previous object pixels with the depth of the shelf back wall at that location. $S_{t+1}^3$ is generated by again updating the depth observation at $S_{t}$ through reprojecting the moved object into the image and replacing the previous object pixels with an object halfway between the moved object depth and back wall.

The policy starts from the current observation $S_t$ as the tree root node. At each search step $k$, the policy samples an action from the generated action set and then samples one of the three resulting states according to the probabilities defined in Eqns.~(\ref{eq:state1})--(\ref{eq:state3}). The policy continues sampling actions and states until it reaches the maximum exploration depth $k_\textrm{max}$ or the branch terminates by sampling $S_{t+k}^1$. We refer to the complete exploration of one branch as a rollout. For each rollout, the reward is $R = \gamma^d \vmathbb{1}_{S_{t+k}^1}(S_{t+k})$, where $S_{t+k}$ is the state of the last node, $\gamma \in [0, 1]$ is a discount factor, and $\vmathbb{1}_{S_{t+k}^1}$ is equal to 1 if the state $S_{t+k}^1$ is sampled at depth $k$ and 0 otherwise. The policy then calculates the score of each action $\mathbf{a}_t$ of the root node as the sum of all rewards at descendants of that action and ranks actions in decreasing order according to their scores, executing the highest ranked feasible action. 
We denote an MCTSSS policy with a tree of maximum depth $k_\mathrm{max}$ and number of total rollouts $M$ as MCTSSS($k_\mathrm{max}$, $M$).




\begin{figure}[t]
    \centering
    \includegraphics[width=0.6\columnwidth]{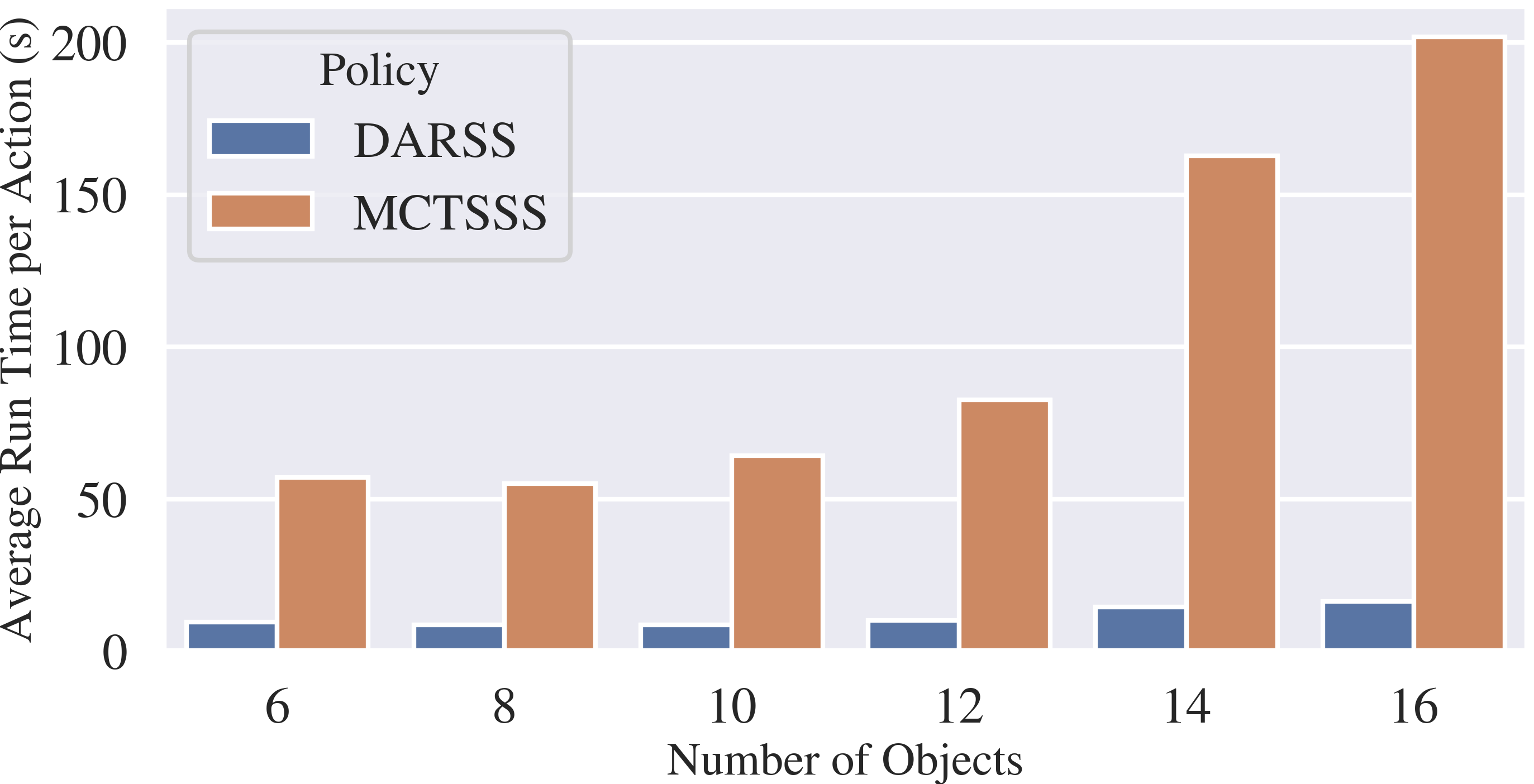}
    \caption{Average run time per action for DARSS and MCTSSS for different numbers of occluding objects in seconds. DARSS scales much better with higher numbers of objects than MCTSSS, taking 8\% of the computation time of MCTSSS for 16 objects.}
    \label{fig:runtime}
    \vspace{-15pt}
\end{figure}
\section{Experiments} \label{sec:experiments}
The LAX-RAY model is trained on 75000 simulated images generated using the First Order Shelf Simulator (FOSS) from~\citet{huang2020mechanical}, modified to allow
stacked environments. We evaluate DARSS and MCTSSS in both simulated and physical stacked shelf environments using a 0.80\,m wide by 0.50\,m high by 0.50\,m deep shelf. We use a green target object with aspect ratio of 1:1 and size 0.06$\times$0.06$\times$0.06\,m$^3$, and we set the target visibility threshold to $V=80\%$. For MCTSSS, we use $k_\mathrm{max} = 2$ and $M = 500$ to balance performance and computation time. For simulation experiments, we use FOSS and set the maximum number of steps to $H = 2N$, where $N$ is the total number of objects. We use the target segmentation mask to estimate the target object visibility. For physical experiments, we set $H = 20$ regardless of the number of objects and use an HSV color threshold to estimate the target object visibility from the observation. We measure success rate (SR, in \%), steps (median, first and third quartiles), and average run time (seconds) per trial for each policy.

\subsection{Baseline and Oracle Policies}
\subsubsection{Baseline Policy}
We propose a heuristic baseline policy such that, for each step, 
picks the object-action pair that has the largest  segmentation mask overlapping reduction in the post-action position. Formally, let $\mathcal{M}_i$ be the segmentation mask of object $\mathcal{O}_i$ in the set of objects $\{\mathcal{O}_1,\dots,\mathcal{O}_N\}$, and $\hat{\mathcal{M}}_i(a_i)$ be the segmentation mask of this object after executing action $a_i$. We pick the object-action pair by
\begin{equation}
    \argmax_{\mathcal{O}_i \in \{\mathcal{O}_1,\dots,\mathcal{O}_N\}, a_i \in \mathcal{A}_p \cup \mathcal{A}_s} M_i - \hat{\mathcal{M}}_i(a_i) \cap \mathcal{M}_{j \neq i}.
\end{equation}

\subsubsection{Oracle Policy}
We consider an oracle policy that has access to full state information including the target object position. It constructs an optimal sequence of actions to reveal the target using dynamic programming.

\begin{table*}[t]
    \centering
    \begin{adjustbox}{width=\textwidth}
    \begin{tabular}{@{}lccccccc@{}}\toprule
Policy & Metric & 6 objs & 8 objs & 10 objs & 12 objs & 14 objs & 16 objs\\
\midrule
\multirow{2}{*}{Oracle} & \SR & \sr{100}  & \sr{100} & \sr{100}  & \sr{100} & \sr{100}  & \sr{100}  \\
& \SA & \sa{1}{1}{2} &  \sa{1}{1}{2}  & \sa{2}{1}{2}  & \sa{2}{1}{2} & \sa{2}{1}{3}  & \sa{2}{2}{3} \\
\midrule
\midrule

\multirow{2}{*}{Baseline} & \SR & \sr{83}  & \sr{79} & \sr{64}  & \sr{50} & \sr{41}  & \sr{22}  \\
& \SA & \sa{3}{2}{5} &  \sa{4}{2}{7}  & \sa{4}{1}{7.25}  & \sa{4}{2}{10.75} & \sa{3}{2}{5}  & \sa{3.5}{2}{7} \\

\midrule
\multirow{2}{*}{DARSS} & \SR & \sr{99}  & \sr*{100} & \sr*{100}  & \sr{98} & \sr{91}  & \sr*{88}  \\
& \SA & \sa{2}{1}{3} &  \sa{3}{2}{4}  & \sa{4}{2}{6}  & \sa{4}{2}{6} & \sa{4}{2}{7}  & \sa{4.5}{3}{7.25} \\

\midrule
\multirow{2}{*}{MCTSSS} & \SR & \sr{99}  & \sr{99} & \sr{99}  & \sr*{100} & \sr*{94}  & \sr{82}  \\
& \SA & \sa{2}{1}{3} &  \sa{3}{2}{4}  & \sa{4}{2}{7}  & \sa{4.4}{3}{7} & \sa{5}{2}{9.75}  & \sa{6.5}{3}{10.75} \\

\bottomrule
    \end{tabular}
    \end{adjustbox}
    \vspace{3pt}
    \caption{DARSS, MCTSSS(2,\,500), Baseline and Oracle policies' success rate (SR) and steps taken, shown as median (first quartile -- third quartile), for 2400 total trials on shelves with 6--16 occluding objects. DARSS and MCTSSS(2,\,500) have similar success rates across different numbers of occluding objects, but DARSS reveals the target object more quickly in shelves with more occluding objects. Both policies outperform the baseline policy and achieve comparable success rate with the oracle policy. Since oracle policy has full state information, it takes fewer steps to find the target object.}
    \label{tab:results}
    \vspace{-25pt}
\end{table*}

\begin{table}[ht]
    \centering
    
\begin{tabular*}{\textwidth}{@{\extracolsep{\fill}}lccccccc@{}}
\toprule
\multirow{2}{*}{Policy}
          & \multicolumn{7}{c}{Number of Occluding Objects} \\
          & 6   & 8   & 10  & 12  & 14  & 16  & 20 \\ \midrule
DAR~\cite{huang2020mechanical} & 78  & 78  & 75  & 56  & 60  & 45  & 34 \\
DAR (destacked) & \textbf{100} & 99  & 98  & 90  & 74  & 55  & 29 \\
DARSS ($-$destack) & 78 & 77 & 77 & 56 & 60 & 49 & 39 \\
DARSS ($-$stack) & \textbf{100} & 99  & 99  & 97  & 85  & 73  & 46 \\
DARSS & 99  & \textbf{100} & \textbf{100} & \textbf{98}  & \textbf{91}  & \textbf{88}  & \textbf{66} \\
\bottomrule
\end{tabular*}
\vspace{3pt}
\caption{Ablation study demonstrating the benefits of stacking and destacking actions for DARSS via success rate (\%) across 3500 total simulated shelves with 6--20 occluding objects. DAR~(destacked) is a DAR policy run on a scene that is first preprocessed by destacking all object stacks and ($-$destack) and ($-$stack) indicate the absence of destacking and stacking actions. DAR~\cite{huang2020mechanical} and DARSS~($-$destack) fail to reveal the targets when object stacks occlude the targets, while DARSS~($-$stack) fails to create space for rearrangement in environments with many objects. While success rates for all policies decline with more occluding objects, these results suggest that both stacking and destacking actions improve policy performance.}
\label{tab:ablation_destack}
\vspace{-25pt}
\end{table}

\subsection{Simulated Experiments}
We compare DARSS, MCTSSS, Baseline, and Oracle policies on a set of 600 shelves with stacked objects. We randomly generate 100 environments each for 6, 8, 10, 12, 14, and 16 occluding objects with the target fully occluded, modified to allow stacked environments. The results are in Table~\ref{tab:results}. The run time in Figure~\ref{fig:runtime} is the average per action using a 72-core Intel Xeon @ 2\,GHz CPUs and 4 Nvidia T4 GPUs parallelized across 8 processes on Google Cloud Platform.

 Without using the occupancy distribution, the Baseline policy does not consider the camera perspective effect and does not encode the search history, resulting in a preference toward moving larger objects, generating repeated actions and the ignorance when the target object is partially occluded. Thus, it has a lower success rate as compared to the other methods. Prior work~\cite{huang2020mechanical} also demonstrates the importance of the occupancy distribution and history encoding.

The multi-step lookahead of MCTSSS allows it to take actions that consider potential future actions; an example is shown in the supplementary material where MCTSSS correctly reveals the target object, while DARSS fails.\jeffi{Passive ``is generated'' here.  Also there seems to be an implication that something is generating the high-quality actions.  Can you clarify?  If the actions are high quality by accident, design, or just likely when due to the nature of the setup?} However, DARSS outperforms MCTSSS when a small set of high-quality actions is generated during the sampling process. The results in Table~\ref{tab:results} suggest that MCTSSS consistently requires a larger number of steps than DARSS while succeeding at a similar rate. MCTSSS suffers from the approximation error of the state sampling probability and prediction error of depth images. These errors propagate as the number of objects increases, which explains the drop in success rate. MCTSSS is also significantly slower than DARSS as shown in Figure~\ref{fig:runtime}, as a depth image of the shelf environment must be generated for each tree state. This state representation results in computation time for MCTSSS growing prohibitively with tree depth and number of rollouts.

Overall, DARSS surprisingly performs on par with or better than MCTSSS despite its lack of lookahead. We hypothesize that since DARSS prioritizes placing objects in regions that do not overlap with the occupancy distribution, it implicitly chooses actions that do not hinder future actions. Both DARSS and MCTSSS achieve a similar success rate compared to the oracle policy but take more steps to reveal the target. Since the Oracle policy has full state information including the target position, no steps are needed for exploration. 

\subsection{DARSS Failure Mode}
The main failure mode of DARSS is \emph{multi-blockage}.
Once \algabbr successfully uncovers a partial view of the target object, the 1D target occupancy distribution collapses to contain only the target object. But the episode will not terminate if the target visibility is lower than the threshold. When a row of objects block the target, the front-most object must be moved first to allow moving other objects in the row to reveal the target. However, the front-most object has minimal overlap with the occupancy distribution due to the perspective effect. Since DARSS is a greedy policy, it chooses actions that immediately decrease the support and acts on the noise in the occupancy distribution in this case. Thus, the policy fails to clear the blockage. An example is shown in the supplementary material.

\subsection{Ablation Experiments}
We measure the effects of stacking and destacking actions by comparing DARSS with ablations that do not have access to stacking or destacking actions. 

DAR~\cite{huang2020mechanical} does not have access to stacking or destacking actions (only pushing and rearrangement actions); thus, DAR can never reveal the target object if it lies behind a stack. Therefore, we also run DAR on \emph{preprocessed} shelf environments, where all stacks are destacked until 1) there are no stacks remaining, or 2) there is insufficient shelf space for more destacking actions. We label this policy as DAR (destacked). Additionally, we include ablations DARSS ($-$stack) and DARSS ($-$destack), which do not have access to stacking and destacking actions, respectively. Table~\ref{tab:ablation_destack} shows the results.

For shelves with less than 12 objects, all policies with destacking actions, as well as DAR (destacked), have nearly 100\,\% success rate, since there is ample space to place and rearrange objects on the shelf. Policies that are unable to destack have over 20\,\% lower success rates, suggesting that destacking actions are critical since the target is often located behind a stack of objects. As the number of occluding objects increases, the preprocessing step used in DAR (destacked) no longer helps. This result suggests that for dense scenes with a large number of objects, \textit{maintaining} stacks in the scenes---either by not destacking all objects or by restacking scenes to clear the shelf floor---is critical for success.
When shelves contain more than 12 objects, the success rate of DARSS ($-$stack) degrades more quickly than that of DARSS as the number of objects increases as there may be insufficient space for rearrangement actions (see supplementary material for examples).
This result suggests that as shelves become more densely packed, the ability to restack objects is critical for clearing shelf space to enable further exploration and rearrangement actions. 

\subsection{Sensitivity Experiments} 
We measure the sensitivity of DARSS to values of $V$ and different target objects with aspect ratios of 2:1 and 4:1. We observe negligible differences in success rates and the median steps required to reveal the target objects for values of $V \in \lbrace 70\,\%, 80\,\%, 90\,\%\rbrace$. For different target objects, we regenerate the set of environments because taller target objects may not be occluded in the same environments where shorter target objects are occluded. DARSS is able to reveal all targets reliably with at least an 88\,\% success rate. Though the difference is small, the median number of steps taken to reveal the target object decreases as the target object aspect ratio increases. Since taller target objects are easier to observe and therefore have fewer fully occluded candidate poses within the shelf, there are fewer locations to be explored and the occupancy distribution is concentrated around fewer objects. Tables with detailed results for these experiments can be found in the supplementary material.

\begin{table}[t]
  \begin{minipage}[b]{0.46\linewidth}
    \centering
   \includegraphics[width=\linewidth]{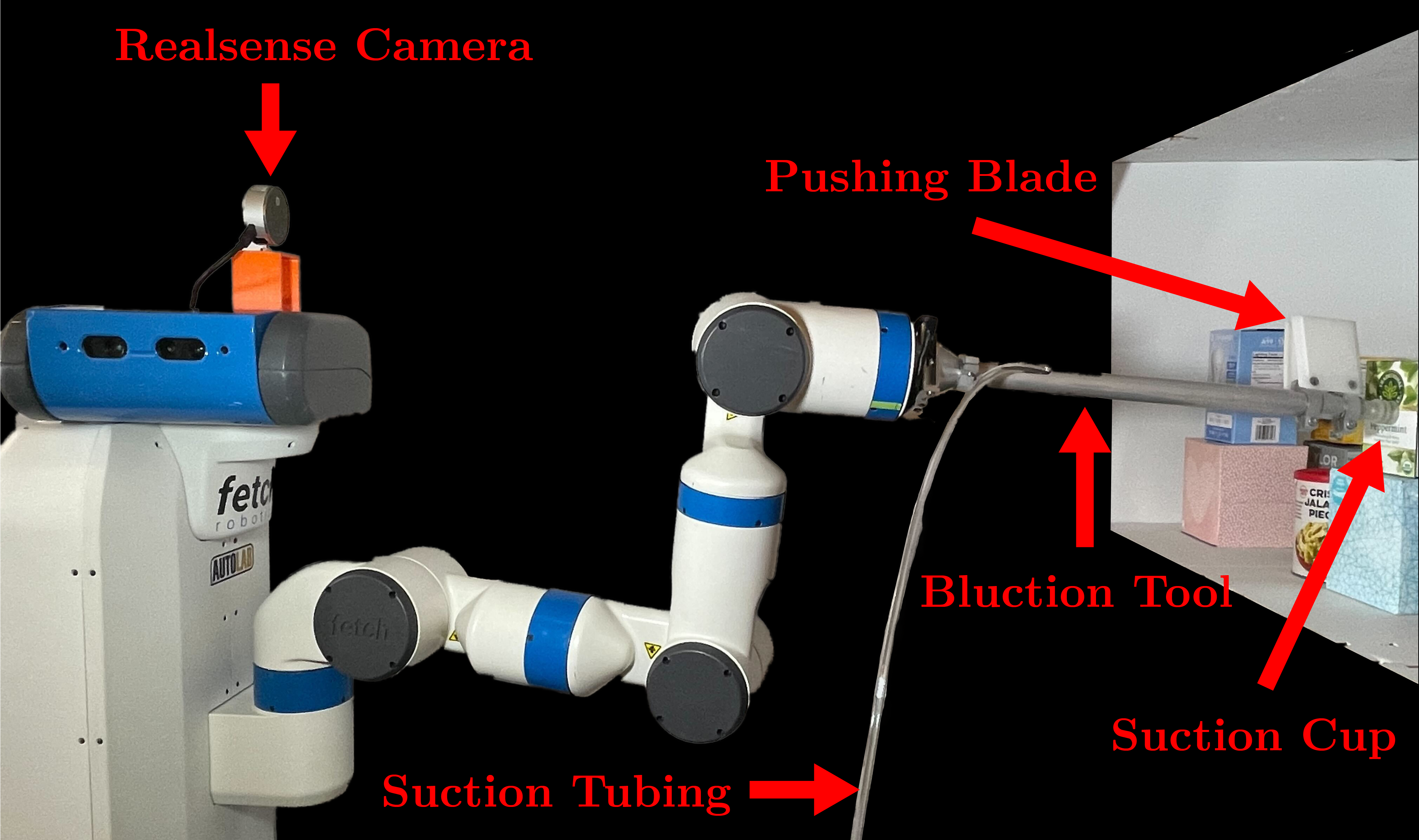}
    \captionof{figure}{Physical experiment setup using a Fetch robot with a bluction tool for mechanical search with stacked objects. Results are shown in Table~\ref{tab:phy}.}
    \label{fig:physs}
  \end{minipage}
  \hfill
  \begin{varwidth}[b]{0.50\linewidth}
  \centering
    \begin{tabular}{@{}lc@{\quad}ccc@{}}
        \toprule
        Policy & Metric & 6 objs & 10 objs & 14 objs \\
        \midrule
        \multirow{2}{*}{\algabbr} & \SR& \srStar{100} & \srStar{100} & \sr{67} \\
        &Steps & 4,\,2,\,2 & 8,\,3,\,6  & 3,\,4,\,F \\
        \midrule
        \multirow{2}{*}{MCTSSS} & \SR& \srStar{100} & \srStar{100} & \srStar{100} \\
        &Steps & 2,\,3,\,2 & 14,\,5,\,16  & 3,\,7,\,5 \\
        \bottomrule
    \end{tabular}
    \vspace{3pt}
    \caption{Physical experiments using a Fetch robot with 3 trials with 6, 10, 14 objects for DARSS and MCTSSS(2,500). Number of steps to find the target object for each experiment are shown. `F' means the policy fails to find the object in 20 steps.}
    \label{tab:phy}
  \end{varwidth}
  \vspace{-25pt}
\end{table}

\subsection{Physical Experiments}
To evaluate the noise robustness of the two proposed policies, we conduct physical experiments using a Fetch robot arm with an attached bluction tool shown in Figure~\ref{fig:physs}. We use an Intel RealSense LiDAR L515 camera to take the RGBD observations.
We use an SD Mask R-CNN~\cite{danielczuk2019segmenting} trained with simulated depth images to generate segmentation masks and use MoveIt!~\cite{moveit} to plan collision free trajectories for the robot arm. 

During the experiments, we assume all the objects can be lifted by the bluction tool and are within the reachable workspace. Since our policies do not consider robot motion planning, as it is outside the scope of the paper, if an action fails due to robot kinematic limitations, we manually set the arm back to an initial position and execute the same action. Our experiments use 6, 10 and 14 objects, each with three different layouts.
An experiment is considered a success when at least 80\% of the target object is revealed within the maximum of 20 steps, and a failure otherwise. 
The dimensions of the shelf and the green target object are the same as in the simulated rollouts. 

Table~\ref{tab:phy} shows results for DARSS and MCTSSS policies. MCTSSS succeeds in all 9 experiments while DARSS fails when the occupancy prediction becomes inaccurate. MCTSSS samples possible future states, which makes it more robust to noisy occupancy distribution estimation than DARSS.  Among all 18 experiments, DARSS requires fewer steps, consistent with simulation experiments.

\section{Conclusion} 
\label{sec:conclusion}
We present a novel mechanical search problem in lateral access environments with stacked objects and two mechanical search policies, DARSS and MCTSSS, that can efficiently reveal a target object. We generate destacking and restacking actions and use the bluction tool~\cite{huang2022mechanical} within the continuous state space. Ablation studies that disallow destacking or restacking actions demonstrate the importance of these two new actions in crowded shelf environments. While simulated experiments show that DARSS outperforms MCTSSS in terms of computation time and the number of steps needed to find the target object, physical experiments suggest MCTSSS may be more robust to perception noise. In this work, we do not use large numbers of rollouts for MCTSSS due to limited computation. For future work, we will focus on increasing the speed and efficiency of MCTSSS and extending both policies to use a continuous action space. 

\printbibliography

\end{document}